\documentclass{article}

\usepackage[preprint]{neurips_2019}
\usepackage{microtype}
\usepackage{graphicx}
\usepackage{subfigure}
\usepackage{booktabs} % for professional tables
\usepackage[dvipsnames]{xcolor}
\usepackage{tikz}
\usepackage{pgfplots}
\pgfplotsset{compat=1.14}
\usepgfplotslibrary{fillbetween}

\usepackage{hyperref}

\title{NASIB: Neural Architecture Search withIn Budget}

\author{%
  Abhishek Singh\thanks{Work done while at Cisco}\hspace{5pt}\thanks{Equal contribution}\\
  MIT Media Labs\\
  \texttt{abhi24@mit.edu}\\
  \And
  Anubhav Garg\footnotemark[2]\\
  Cisco Systems\\
  \texttt{anubhgar@cisco.com}\\
  \AND
  Jinan Zhou\\
  Cisco Systems\\
  \texttt{jinazhou@cisco.com}\\
  \And
  Shiv Ram Dubey\\
  IIIT Sri City\\
  \texttt{srdubey@iiits.in}\\
  \And
  Debo Dutta\\
  Cisco Systems\\
  \texttt{dedutta@cisco.com}\\
}

\begin{document}

\maketitle

\begin{abstract}
Neural Architecture Search (NAS) represents a class of methods to generate the optimal neural network architecture, and typically iterate over candidate architectures till convergence over some particular metric like validation loss. They are constrained by the available computation resources, especially in enterprise environments. In this paper, we propose a new approach for NAS, called NASIB, which adapts and attunes to the computation resources (budget) available by varying the exploration vs exploitation trade-off. We reduce the expert bias by searching over an augmented search space induced by {\em Superkernels}. The proposed method can provide the architecture search useful for different computation resources and different domains beyond image classification of natural images where we lack bespoke architecture motifs and domain expertise. We show, on CIFAR10, that it is possible to search over a space that comprises of 12x more candidate operations than the traditional prior art in just 1.5 GPU days, while reaching close to state of the art accuracy. While our method searches over an exponentially larger search space, it could lead to novel architectures that require lesser domain expertise, compared to the majority of the existing methods.
\end{abstract}

\section{Introduction}

Neural Architecture Search (NAS) embodies a wide variety of techniques to yield the optimal neural network architecture based on a variety of constraints while optimizing for one or more objectives (\cite{nasnet,metaqnn,mnas,ppp-net,plnas}). This helps to avoid the arduous manual tuning of models and has gained a lot of attraction in the last few years and has been applied to a variety of scenarios, including generating compact models for mobile devices, etc. We believe that NAS will accelerate the adoption of deep learning architectures in the industry by making it simpler to optimize such architectures for a wide variety of use cases that run on heterogeneous platforms (e.g. mobile, IoT, etc.). 

Current works in the NAS space tend to have the following two limitations:
\begin{enumerate}
    \item The search method is designed with the assumption that the user will run it till  convergence. It is not clear how to leverage these methods  when the computational budget for performing architecture search is significantly lower or higher than that used by such methods.
    \item The search space used is quite narrow (relatively) and is built on already optimal structural decisions, hence does not provide any significant benefit over a random search.
\end{enumerate}

In this paper, we propose a simple, intuitive and effective solution to both the problems using a common framework, called {\em NASIB} and present a trade-off between the computational budget and exploration over a vast space of neural network architectures. Unlike the traditional NAS frameworks which start with a pre-filtered fixed search space (where the filtering is done by domain experts), we start with an unbiased search space by considering a huge space of various possible candidate operations (e.g.,  for CNNs we search over all possible kernels with all combinations of kernel height and width less than $10$, which amounts to \textbf{81} different kernels at every layer) and dynamically reduce the likelihood of the ineffective operations. To ensure the efficiency of our framework under such a huge set of structural decisions, we adapt to the computational budget available for search by reducing the \textit{rate} of dynamic pruning of the search space since such pruning rates affect the amount of resources consumed, a proxy for the cost. 
This adaptive pruning is the trade-off between exploration and exploitation over a huge search space of architectures.

Most prior works focus on designing network architectures with high accuracy and efficiency, but 1) do not use available computational resources as a constraint and 2) use a smaller search space based on the prior knowledge about the task at hand, e.g., previously used successful architectures. This has led to the design of search spaces where every possible structural decision itself is close to the optimal architecture. This is evident from the insignificant difference in the performance of architectures discovered by well known architecture search methods against their random baselines (\cite{amoebanet,darts,oneshot}).

\begin{figure}
    \centering
    \includegraphics[width=0.95\textwidth,height=2.7cm]{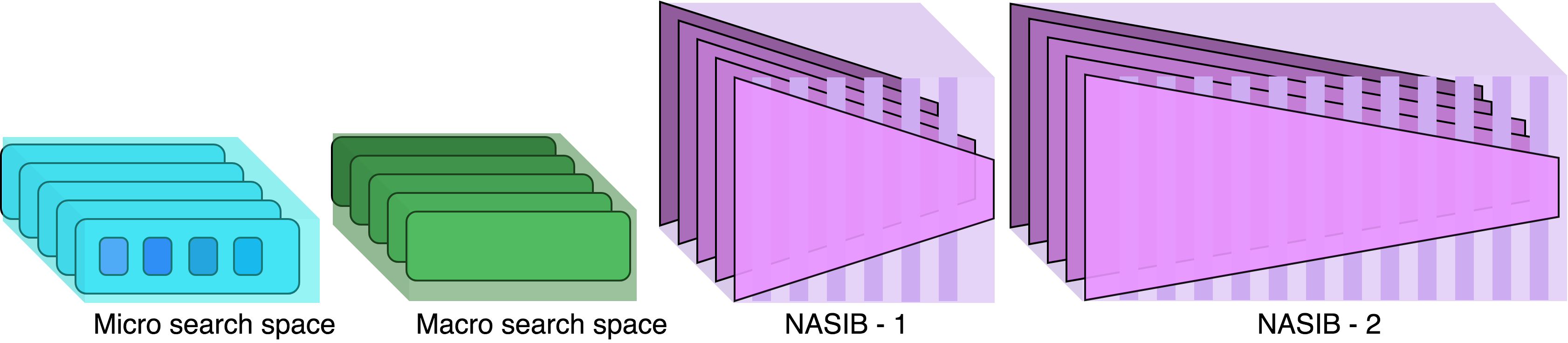}
    \caption{Juxtaposition of our method with traditional NAS methods. The first diagram refers to the micro search space in the architecture search where the topology of operations is searched to construct a single block and the same single block is stacked multiple times to obtain the final architecture. The next diagram is for the macro search space where candidate operations are searched at every layer. Similar to the macro search space, NASIB also searches over the candidate operations at every layer. However, probabilistically, it reduces a relatively bigger search space to a smaller one based on their relative performance. NASIB-1 illustrate the NAS workload with computation budget similar to the previous two search paradigms. NASIB-2 illustrates the search workload with higher computation budget.}
    \label{fig:search_space}
\end{figure}

In this paper, we propose an architecture design method which searches over an exponentially larger search space, requires lesser domain expertise, and presents an  algorithm that leverages the exploration vs exploitation trade-off and automatically adapts over the computation resources available for performing the search. The benefit of proposed budget aware NASIB neural architecture search method is illustrated in Fig. \ref{fig:search_space} against the traditional NAS schemes such as Micro and Macro search space. The NASIB-1 and NASIB-2 demonstrate the search space for the low and high computation budget available, respectively. We empirically demonstrate superiority of the proposed method over existing methods by searching over a space having roughly \textbf{12X} times more candidate operations than the majority of the recent search methods under same computation budget without any compromise on the performance of the discovered architecture. 

\subsection{Our Contributions}
In this paper, we have built a novel approach to NAS that could pave the way for architecture which requires less domain expertise while optimizing for budget constraints. Some of our key contributions are as follows: 
\begin{itemize}
    \item We take the first steps towards designing a gradient based neural architecture search method which optimizes the search complexity for the computation budget available without compromising the performance of the discovered architectures. The proposed method also provides ease of usability to the end user by simplifying the computation budget specification to just hours or days.
    \item We design the search space of CNN architectures by introducing \textit{superkernels}, which is a big set of candidate operations. By using this increased candidate operations, we increase the search space by \textbf{12X} more candidate operations at every layer in the architecture.
    \item We show that the well known method like ENAS (\cite{enas}) can be further optimized by using a bigger and diverse search space. Just by augmenting the search space of ENAS, we increase the validation accuracy of architecture search by a margin of \textbf{10\%}.
    \item We perform the extensive experiments using the proposed method by varying the computation budgets to show its efficacy over CIFAR10 dataset. Moreover, we also show the relative efficiency of our method by comparing the relative performance during the search.
\end{itemize}
The rest of the paper is structured in the following way: Section~\ref{sec:related} presents the related work. Then we introduce the budget aware NASIB approach in Section~\ref{sec:nasib} followed by experiments and results in Section~\ref{sec:results}. We then present some observations in Section~\ref{sec:discussion} and conclude in Section~\ref{sec:conclusion}. 
\section{Related Work}
\label{sec:related}
Designing  neural network architectures have been studied for a long time (\cite{old_1,old_2,old_3,old_4}). Recent results showing the efficacy of reinforcement learning based methods (\cite{metaqnn,nasnet}) to design neural network architectures has sparked off research in this field. Since then, a variety of methods have been proposed which can be grouped into three categories - a) New techniques for performing the architecture search, b) improving the efficiency of the search, and c) applying architecture search to different domains or performing multi-objective architecture search. Our work is orthogonal to all of the above mentioned three categories, and focuses on performing the search based on the computation resources available for the user. Since we make this search over a big search space of candidate operations, this work has intersection with the category (c). We also reduce the expert bias in the design of architectures. 

NAS methods include reinforcement learning (\cite{nasnet,metaqnn,network_transformation,irlas}), evolution (\cite{amoebanet, lrg_scale_evo, lamarckian_evo}), and gradient based methods (\cite{das, darts, snas, plnas}) which are widely used. In \cite{reinfo-evolution}, the authors combine using reinforcement learning to guide the mutation process of evolutionary algorithms to evolve neural network architectures. There are several other methods which are based upon Bayesian Optimization (\cite{kandasamy,netmorphism}), Auto-Encoder (\cite{nao}), network morphism (\cite{netmorphism}), pruning (\cite{convfabrics, nacnet, anotherpruning}), and etc. \cite{survey} and \cite{new_survey} provide a comprehensive summary of various methods developed for NAS.

For efficiency, ENAS (\cite{enas}) and performance prediction (\cite{perfprediction}) improve the search time of NASnet (\cite{nasnet}) and MetaQNN (\cite{metaqnn}), respectively. One-shot model hypernetworks (SMASH) (\cite{smash}) and Graph Hypernetworks (\cite{ghypernetworks}) learn the models which can generate the weights for a target model without training. One shot architecture search (\cite{oneshot}) trains overprovisioned network similar to ENAS (\cite{enas}) and samples a sparse path during the validation phase.
Recent work in multi-objective NAS (\cite{lamarckian_evo}, \cite{mnas}, \cite{platformaware}, \cite{instanas}, \cite{ppp-net}) focuses on discovering models satisfying multiple constraints in addition to accuracy like latency, number of parameters, multi-add operations, and etc.
All previous \textit{hardware aware} architecture search methods refer to the hardware awareness of the model discovered from search while our work, to the best of our knowledge, is the first attempt to have an architecture search method which is aware of the computation resources available. 
\section{NASIB}
\label{sec:nasib}
In this section, we present our approach, NASIB, an adaptive NAS method that adapts to the available computation budget.
\begin{figure}
    %\centering
    \begin{minipage}{.45\textwidth}
    \centering
    \begin{tikzpicture}[]
    \centering
    \begin{axis}[
            xlabel={Epochs},
            ylabel={Validation Accuracy},
            grid style=dashed,
            xmin=-3,
            legend style={legend pos=south east,font=\scriptsize},
            height=1.\textwidth,
            width=1.\textwidth,
        ]
        \addplot[
            color=green,
            ]
        table {data/enase_acc_compairson_mean_big_ss.dat};
        \addlegendentry{ENAS on bigger search space}
        \addplot[
            color = orange,
            ]
        table {data/enase_acc_compairson_mean_orig_ss.dat};
        \addlegendentry{Original ENAS}
        \addplot[
            name path=std_high_1,
            color=green!50
            ]
        table {data/enase_acc_compairson_high_big_ss.dat};
        \addplot[
            name path=std_low_1,
            color=green!50
            ]
        table {data/enase_acc_compairson_low_big_ss.dat};
        \addplot[
            name path=std_high_2,
            color=orange!50
            ]
        table {data/enase_acc_compairson_high_orig_ss.dat};
        \addplot[
            name path=std_low_2,
            color=orange!50
            ]
        table {data/enase_acc_compairson_low_orig_ss.dat};
        \addplot[green!30,fill opacity=0.5] fill between[of=std_high_1 and std_low_1];
        \addplot[orange!30,fill opacity=0.5] fill between[of=std_high_2 and std_low_2];
    \end{axis}
    \end{tikzpicture}
    \caption{Mean accuracy with standard deviation for the two set of experiments on the ENAS implementation. The green curve corresponds to the accuracy obtained with a bigger search space and the orange curve corresponds to the running of the original ENAS. The search space here is for the macro search space and the dataset is CIFAR10.}
    \label{enasvsbigenas}
    \end{minipage}%
    \hspace{1.1cm}
    \begin{minipage}{.45\textwidth}
    \centering
    \begin{tikzpicture}[]
    \centering
    \begin{axis}[
            xlabel={Epochs},
            ylabel={Validation Accuracy},
            grid style=dashed,
            xmin=-2,
            xmax=200,
            height=1.\textwidth,
            width=1.\textwidth,
            legend style={legend pos=south east, font=\scriptsize},
        ]
        \addplot[
            color=blue,
            each nth point={8},
            mark=diamond
            ]
        table {data/comparison_other_methods/darts.dat};
        \addlegendentry{DARTS}
        \addplot[
            color=brown,
            each nth point={8},
            mark=*
            ]
        table {data/comparison_other_methods/enas.dat};
        \addlegendentry{ENAS}
        \addplot[
            color=red,
            each nth point={8},
            mark=triangle
            ]
        table {data/comparison_other_methods/snas.dat};
        \addlegendentry{SNAS}
        \addplot[
            color=purple,
            each nth point={8},
            mark=square
            ]
        table {data/comparison_other_methods/nasib.dat};
        \addlegendentry{NASIB-1}
        \addplot[
            color=RedViolet,
            each nth point={8},
            mark=o
            ]
        table {data/comparison_other_methods/nasib_superkernel.dat};
        \addlegendentry{NASIB-2}
    \end{axis}
    \end{tikzpicture}
    \caption{Comparison of validation accuracy of different methods during the architecture search process. NASIB-1 refers to the architecture search performed using the widely used narrow search space and NASIB-2 refers to the search performed using superkernels. All other search method belong to the micro search space category.}
    \label{searchcomparison}
    \end{minipage}
\end{figure}
\label{method}
\subsection{From Computation Resources to Epochs}
One key aspect of this work is to map available computation resources in a quantifiable manner which can be utilized by the search algorithm. In the current state, the computation resources for a NAS workload are compared in terms of GPU days, which essentially means the number of GPUs utilized multiplied by the number of days required to perform the search. This approach of benchmarking is inconsistent due to a variety of reasons. First, for any deep learning workload, there are a myriad of factors, dependent as well as independent of the GPUs used, which contribute to the processing speed of every pass of forward and backward propagation. Different GPUs exhibit significantly different computation speeds. Furthermore, CPU speed, memory access rate and many more interwoven components are also a key factor in deciding the overall speed. Therefore, for our search algorithm to adapt to the available computation resources, a comprehensive list of all these components and their specifications would be required. However, from an end user's perspective, it is far more convenient to simply specify only the number of hours/days and give the whole system/cluster as a black box to the search algorithm. Hence, rather than developing a complex mapping of computation resources, we instead take the number of hours/days available for search as an input and divide it by the time required by a single epoch to complete on the provided system. This gives the total number of epochs available for performing the search. This way of mapping available computation resources to the number of epochs is simple and yet provides a most consistent and accurate approximation of computation resources available because we simply view the system as a black box which takes all possible interwoven constituents into the account when measuring the time required for a single epoch. This approach is also free from any function approximator of hardware configurations and hence is expected to work across a wide range of hardware configurations ranging from embedded devices to big multi-GPU clusters.

\subsection{Differentiable Architecture Search}
Our architecture search technique is built upon the existing differentiable architecture search methods (\cite{darts,plnas,snas}). We start with a base over-parameterized network comprising of all candidate operations and search for a smaller sub-network. In order to obtain this sub-network, all nodes are assigned a single scalar $\alpha$ also called architecture parameter. The $\alpha$ is kept trainable and updated via gradient based methods. After the whole search method is over, the node with highest $\alpha$ for a given layer is retained and rest all are pruned away. This way, after training we are left with a compact architecture obtained by pruning majority of the nodes from the overprovisioned base network. Similar to the scheme used by ProxylessNAS (~\cite{plnas}), we sample only two operations for every single batch instead of $N$ operations in order to reduce the memory requirements in the GPU. Sampling of every candidate operation at any given layer is dependent upon the magnitude of their respective $\alpha$. This sampling procedure is indispensable for our framework because of the huge set of candidate operations introduced by superkernels used at every layer. The training is performed similar to the method proposed in DARTS (~\cite{darts}. The parameters of the operations are trained based on eq.~\ref{eq1} and in the next round architecture parameters, $\alpha$'s are trained based on eq.~\ref{eq2} while treating the network parameters as constant.
\noindent\begin{minipage}{.5\linewidth}
\begin{equation}
  \label{eq1}
  x_{\ell} = \sum_{j=1}^{N}O_{\ell,j}(x_{\ell-1})
\end{equation}
\end{minipage}%
\begin{minipage}{.5\linewidth}
\begin{equation}
  \label{eq2}
  x_{\ell} = \sum_{j=1}^{N}{\frac{\exp(\alpha_{j}^{\ell})}{\sum^{N}_{i=1}\exp(\alpha_{i}^{\ell})}O_{\ell,j}(x_{\ell-1})}
\end{equation}
\end{minipage}
%\iffalse
%\fi
Here, $O_{\ell,j}$ is the $j^{th}$ candidate operation at the layer $\ell$, $x_{\ell}$ is the output of the layer $\ell$ and $N$ is the total number of candidate operations. One noteworthy advantage of differentiable method is the credit assignment to every individual node in the search space by architecture parameter $\alpha$ whose relative magnitude act as a proxy for credit assignment over structural decisions. This is different from some of the other commonly used approaches in architecture search like reinforcement learning and evolutionary methods where the complete architecture receives a score.
\subsection{Policy based Sampling from Search Space}
\begin{figure}[t!]
    %\centering
    %\resizebox{10cm}{2cm}{%
    \begin{tikzpicture}[]
    \begin{axis}[
            width=1.02\textwidth,
            height=0.4\textwidth,
            xlabel={Filters},
            ylabel={Frequency},
            grid style=dashed,
            xtick distance=1,
            xmin=0.8,
            xmax=18.5,
            ticklabel style = {font=\tiny},
            xticklabels={3,3,3x3,3sep,5x5,5x5sep,avgpool,maxpool,2x2,4x4,6x6,2x5,6x3,4x9,9x3,9x5,1x6,3x4,4x10,10x5},
            legend style={legend pos=north west,font=\tiny},
        ]
        \addplot[
            color=red
            ]
        table {data/filter_type_enas_data/enas_filters_s1_l1.dat};
        \addlegendentry{Experiment set 1}

        \addplot[
            color=blue
            ]
        table {data/filter_type_enas_data/enas_filters_s2_l1.dat};
        \addlegendentry{Experiment set 2}

        \addplot[
            color=yellow
            ]
        table {data/filter_type_enas_data/enas_filters_s3_l10.dat};
        \addlegendentry{Experiment set 3}
        
		\addplot[
            color=red
            ]
        table {data/filter_type_enas_data/enas_filters_s1_l2.dat};

		\addplot[
            color=red
            ]
        table {data/filter_type_enas_data/enas_filters_s1_l3.dat};

		\addplot[
            color=red
            ]
        table {data/filter_type_enas_data/enas_filters_s1_l4.dat};

		\addplot[
            color=red
            ]
        table {data/filter_type_enas_data/enas_filters_s1_l5.dat};

		\addplot[
            color=red
            ]
        table {data/filter_type_enas_data/enas_filters_s1_l6.dat};

		\addplot[
            color=red
            ]
        table {data/filter_type_enas_data/enas_filters_s1_l7.dat};

		\addplot[
            color=red
            ]
        table {data/filter_type_enas_data/enas_filters_s1_l8.dat};

		\addplot[
            color=red
            ]
        table {data/filter_type_enas_data/enas_filters_s1_l9.dat};

		\addplot[
            color=red
            ]
        table {data/filter_type_enas_data/enas_filters_s1_l10.dat};

		\addplot[
            color=red
            ]
        table {data/filter_type_enas_data/enas_filters_s1_l11.dat};

		\addplot[
            color=red
            ]
        table {data/filter_type_enas_data/enas_filters_s1_l12.dat};

		\addplot[
            color=blue
            ]
        table {data/filter_type_enas_data/enas_filters_s2_l2.dat};

		\addplot[
            color=blue
            ]
        table {data/filter_type_enas_data/enas_filters_s2_l3.dat};

		\addplot[
            color=blue
            ]
        table {data/filter_type_enas_data/enas_filters_s2_l4.dat};

		\addplot[
            color=blue
            ]
        table {data/filter_type_enas_data/enas_filters_s2_l5.dat};

		\addplot[
            color=blue
            ]
        table {data/filter_type_enas_data/enas_filters_s2_l6.dat};

		\addplot[
            color=blue
            ]
        table {data/filter_type_enas_data/enas_filters_s2_l7.dat};

		\addplot[
            color=blue
            ]
        table {data/filter_type_enas_data/enas_filters_s2_l8.dat};

		\addplot[
            color=blue
            ]
        table {data/filter_type_enas_data/enas_filters_s2_l9.dat};

		\addplot[
            color=blue
            ]
        table {data/filter_type_enas_data/enas_filters_s2_l10.dat};

		\addplot[
            color=blue
            ]
        table {data/filter_type_enas_data/enas_filters_s2_l11.dat};

		\addplot[
            color=blue
            ]
        table {data/filter_type_enas_data/enas_filters_s2_l12.dat};

		\addplot[
            color=yellow
            ]
        table {data/filter_type_enas_data/enas_filters_s3_l1.dat};

		\addplot[
            color=yellow
            ]
        table {data/filter_type_enas_data/enas_filters_s3_l2.dat};

		\addplot[
            color=yellow
            ]
        table {data/filter_type_enas_data/enas_filters_s3_l3.dat};

		\addplot[
            color=yellow
            ]
        table {data/filter_type_enas_data/enas_filters_s3_l4.dat};

		\addplot[
            color=yellow
            ]
        table {data/filter_type_enas_data/enas_filters_s3_l5.dat};

		\addplot[
            color=yellow
            ]
        table {data/filter_type_enas_data/enas_filters_s3_l6.dat};

		\addplot[
            color=yellow
            ]
        table {data/filter_type_enas_data/enas_filters_s3_l7.dat};

		\addplot[
            color=yellow
            ]
        table {data/filter_type_enas_data/enas_filters_s3_l8.dat};

		\addplot[
            color=yellow
            ]
        table {data/filter_type_enas_data/enas_filters_s3_l9.dat};

		\addplot[
            color=yellow
            ]
        table {data/filter_type_enas_data/enas_filters_s3_l11.dat};

		\addplot[
            color=yellow
            ]
        table {data/filter_type_enas_data/enas_filters_s3_l12.dat};

    \end{axis}
    \end{tikzpicture}
    %}
    \caption{Higher frequency denotes higher confidence displayed by the sampler for a given candidate operation. The first 6 operations are from the original search space and rest 12 operations have been introduced by us while keeping every other parameter and hyperparameter same. While experiments do not seem to prefer the same set of structural decisions over the others, nevertheless it clearly asssigns high confidence to many of the operations from the extended search space. Most of these additional candidate operations are rarely used in the CNN architectures.}
    \label{enasfiltersdiscovered}
\end{figure}
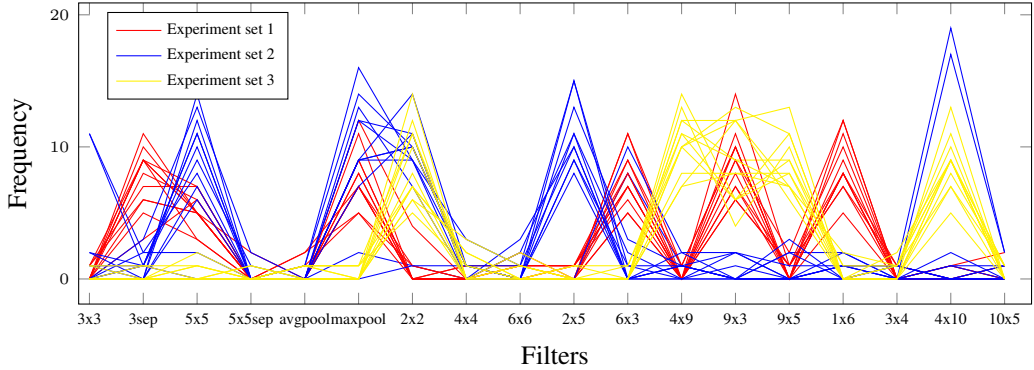

To account for the exploration and exploitation phase during the search, we use two policies $P_1$ and $P_2$, respectively. For exploration, we encourage sampling of those operations whose sampling frequency is relatively lower than the majority of operations. Under the policy $P_1$ and any given layer $l$, the sampling probability $p_{\ell}^{i}$ of a node $i$ is computed by calculating the negative softmax of previously sampled frequencies as per eq.~\ref{eq3},
\noindent\begin{minipage}{.5\linewidth}
\begin{equation}
  \label{eq3}
  p_{\ell}^{i} = \frac{\exp(-freq_{\ell}[i])}{\sum^{N}_{j=1}\exp({-freq_{\ell}[j]})}    
\end{equation}
\end{minipage}%
\begin{minipage}{.5\linewidth}
\begin{equation}
  \label{eq4}
  p_{\ell}^{i} = \frac{\exp(\alpha_{i}^{\ell}/\lambda)}{\sum^{N}_{j=1}\exp(\alpha_{j}^{\ell}/\lambda)}
\end{equation}
\end{minipage}
The negative sign in front of all the sampled frequency numbers allows to assign higher probability to the nodes with lower sampling frequency in the past. Hence, it encourages the exploration of unexplored nodes. The reasoning behind using softmax of frequencies instead of random sampling, which is a commonly used method in reinforcement learning to promote exploration, is to explicitly encourage those operations which have been sampled with low frequency before to encourage the diversity in the architecture search.
For the policy $P_2$, a node is sampled with probability based on the softmax of architecture parameters, $\alpha$ as shown in eq.~\ref{eq4}.
Since the policy $P_2$ is expected to be purely exploitative in nature, we keep the value of $\lambda$ to be very low. While a low value for $\lambda$ ensures the significant hardness in the softmax (high temperature), having $\lambda\to0$ turns the softmax to hardmax, hence prohibiting the sampling of candidate operations with high value of $\alpha$ but less than the maximum alpha which is usually the promising candidate operations. In the policy $P_2$, we sample two candidate operations, which is similar to the setup described in ProxylessNAS (\cite{plnas}). We sample one out of the two policies for every layer independently during the training phase and the policy is resampled after every batch of training data. The policy $P_{1}$ is sampled with probability $p_i$ during every epoch, here $i$ is the i\textsuperscript{th} epoch. We initialize with $p_1 = 1$ during the beginning of the architecture search so as to explore all candidate operations in the beginning. The $p_1$ is exponentially decayed with the decay parameter $\beta$ such that $p_i = 1.\beta^{i}$. We decay the probability $p_i$ in such a way that the value of $p_n=k$ in the last epoch. The value of $k$ can also be viewed as one of the hyperparameter to the proposed search method. We found $k=0.1$ to be working well for our experimental setup. The value of $\beta$ is calculated as $\beta = e^{\frac{\ln{k}}{n}}$ before beginning of the training procedure. As the probability of sampling $P_2$ increases, the sampling probability of nodes with low value of $\alpha$ decreases. This, in essence, means that even though we start with a big search space, the search space gets pruned \textit{softly} as we proceed in the exploitation phase guided by the policy $P_2$.

\subsection{Searching for CNN Architectures}
\label{cnnsearch}
In this section, we discuss the proposed approach to search for the CNN architectures designed for the task of image classification. In order to reduce the expert bias in the design of the search space, we introduce the notion of superkernel. A superkernel of size $m$ can be viewed as a square having size $m$ which encompasses all possible combinations of rectangles of dimensions $i\times j, \forall i,j\leq m$. Thus, a superkernel of size $m$ would result in $m^2$ different rectangles. All such rectangles can be viewed as kernels of dimensions $i\times j$. This superkernel act as a good representative of various possible kernels which can be discovered through the optimization procedure at any given layer in the architecture. Note that this superkernel idea is not similar to the superkernels idea introduced by \cite{DBLP:journals/corr/abs-1904-02877} because they consider only three kernels (i.e., 3, 5, and 7) which are square and have odd length. Also, they use the idea of superkernels to share parameters and learn the kernel parameters function efficiently which is orthogonal to our objective. We use macro search space for architecture search and at every layer we utilize a single superkernel. Therefore, at every layer, we allow all possible candidate operations to be searched over, but at any given time only two operations are sampled from the superkernel which reduces the memory requirement. After the training is over, we only retain the operations with highest $\alpha$'s at every layer to derive the final architecture. We assign equal probability to all candidate operations in our experiments, but allow the user to provide priors on selected candidate operations in order to induce expert knowledge. This would allow discovering desirable architectures faster as the search algorithm without any priors would spend an equal amount of time in exploring all operations initially.
For larger value of $m$ such as $m\geq 5$, the memory requirements to store the parameters of even a single kernel with multiple channels would become very high, hence we fix a threshold $K$ such that kernels whose parameters for a single channel exceed $K$, i.e. $i\times j>K$, are turned into the depthwise separable convolution which is efficient in terms of number of parameters. For our experiments we keep $K=9$.
\section{Experiments and Results}
\label{sec:results}
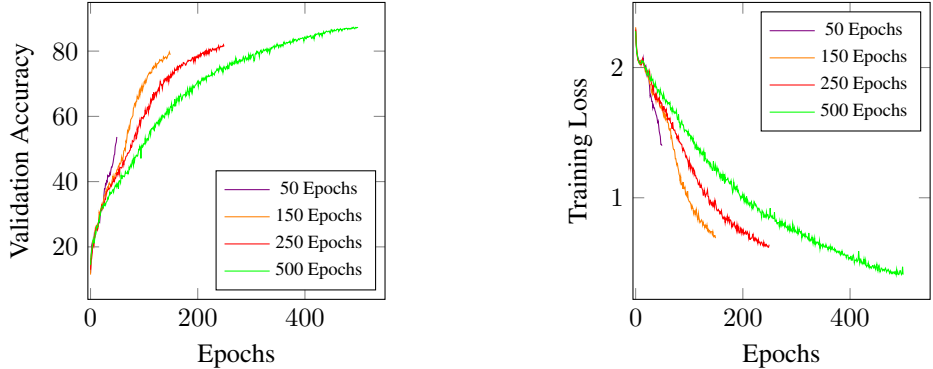
\begin{figure}
    %\centering
    \begin{minipage}{.44\textwidth}
    \centering
    \begin{tikzpicture}[]
    \centering
    \begin{axis}[
            xlabel={Epochs},
            ylabel={Validation Accuracy},
            grid style=dashed,
            xmin=-5,
            legend style={legend pos=south east,font=\scriptsize},
            height=0.9\textwidth,
            width=0.9\textwidth,
        ]
        \addplot[
            color=violet,
            ]
        table {data/adaptive_epoch/50.dat};
        \addlegendentry{50 Epochs}
        \addplot[
            color=orange,
            ]
        table {data/adaptive_epoch/150.dat};
        \addlegendentry{150 Epochs}
        \addplot[
            color=red
            ]
        table {data/adaptive_epoch/250.dat};
        \addlegendentry{250 Epochs}
        \addplot[
            color=green
            ]
        table {data/adaptive_epoch/500.dat};
        \addlegendentry{500 Epochs}
    \end{axis}
    \end{tikzpicture}
    \label{fig:validation_accuracy}
    \end{minipage}%
    \hspace{1.1cm}
    \begin{minipage}{.44\textwidth}
    \centering
    \begin{tikzpicture}[]
    \centering
    \begin{axis}[
            xlabel={Epochs},
            ylabel={Training Loss},
            grid style=dashed,
            xmin=-5,
            height=0.9\textwidth,
            width=0.9\textwidth,
            legend style={legend pos=north east, font=\scriptsize},
        ]
        \addplot[
            color=violet,
            ]
        table {data/rs/epoch50.dat};
        \addlegendentry{50 Epochs}
        \addplot[
            color=orange,
            ]
        table {data/rs/epoch150.dat};
        \addlegendentry{150 Epochs}
        \addplot[
            color=red
            ]
        table {data/rs/epoch250.dat};
        \addlegendentry{250 Epochs}
        \addplot[
            color=green
            ]
        table {data/rs/epoch500.dat};
        \addlegendentry{500 Epochs}
    \end{axis}
    \end{tikzpicture}
    \label{fig:training_loss}
    \end{minipage}
    \caption{Search progress under different computational budget on CIFAR10 dataset with superkernel search space. We keep all hyperparameters same except the computational budget, which is provided in such a way that they get mapped to the mentioned value of target epochs.}
    \label{performnace}
\end{figure}
We perform several sets of experiment to test our hypothesis and demonstrate the efficacy of the proposed method. All experiments are performed on CIFAR10 (\cite{cifar10}), which is a standard dataset for benchmarking NAS workloads on the image classification task. All experiments are performed on a single Nvidia V100 GPU. For testing of our hypothesis on usage of more number of candidate operations, we use the same codebase of ENAS (\cite{enas}) and simply increase the search space by adding more number of convolution filters with varying kernel sizes. All hyperparameters and other experimental details are kept exactly the same as provided in their publicly released codebase for performing macro search on CIFAR10 dataset. As shown in Figure \ref{enasvsbigenas}, there is a significant increase in the accuracy of ENAS with extended search space throughout the validation phase of architecture search. Note that the whole training setting, including hyper-parameters remain same for both the curves except the search space which is larger for the original curve. Although we increase the search space for ENAS in our experiments, we do not observe any increase in the wall clock time between two epochs. The improvement in the performance might indicate that ENAS is able to utilize its computational budget with augmented search space.
Experimentally, it confirms our argument about the usefulness of the diversity in the search space in obtaining high performance. We take this study performed on ENAS one step further by analyzing the architectures sampled from the controller after the search phase is over. In this study we perform the architecture search for $3$ different trials and on the conclusion of each trial we sample $50$ architectures from the controller. For results obtained from each trial, we sum the frequency of convolutional filters sampled to understand the distribution learned by the controller. 
We plot the results in Figure~\ref{enasfiltersdiscovered}. As shown, the distribution of architectures obtained across three different trials clearly indicates that when offered with more choices of kernels in the search space, the controller learned to choose some of the kernels which are not widely used in the majority of the CNN architectures used for image classification.\\

Similar to the study performed in \cite{snas} over the relative performance of architectures during the search phase, we also compare two of our workloads in Figure~\ref{searchcomparison}. As it can be observed, NASIB-1 outperforms other methods when trained with a small search space similar to the other methods, we attribute this to efficient utilization of computational budget. Note that while the curve corresponding to DARTS converges faster than NASIB-1, \cite{snas} showed that there is a strong inconsistency with the performance of its child network.\\

We show the adaptability of our method towards computation resources available for performing the search by searching for architectures on CIFAR10 for four different scenarios. All four scenarios have been formed by allocating different computation budget for performing the search. In figure~\ref{performnace}, we plot validation accuracy as well as the training loss as the search progresses for all four workloads. It can be observed from the plot that slope of the curve decreases as the workloads approach close to their target epochs and this slope varies significantly for all four different workloads which means lower the number of epochs available, the model spends relatively lesser time in exploring the search space and hence converges quickly by entering the exploitation phase. Note that the optimal operator selection takes some fixed amount of time as the algorithm requires some initial time to perform credit assignment to every individual operation which can be done only after training all these operations to a certain extent.
\begin{table}
\small
\centering
\caption{Comparison of performance on two standard benchmarks. Search time is reported in GPU days. $\#OPs$ refers to the number of candidate operations in the search space.}

%\begin{tabular}{|l{1.6cm}|S{1.5cm}|S{1.1cm}|S{1.2cm}|S{1.1cm}|S{1.6cm}}

\begin{tabular}{|l|l|l|l|l|l|}
    \specialrule{.1em}{.05em}{.05em}
    Network & Dataset & Params & Test error & \#OPs& Search time\\
    \specialrule{.1em}{.1em}{.1em} 
    \specialrule{.1em}{.05em}{.05em}
    NASNet-v3 (\cite{nasnet}) & CIFAR10 & 37.4M & 3.65 & 13 & 1800\\
    \hline
    Block-QNN (\cite{metaqnn}) & CIFAR10 & 39.8M & 3.54 & 8 & 96\\
    \hline
    AmoebaNet-B (\cite{amoebanet}) & CIFAR-10 &  34.9M & 2.13 & 19 & 3150\\
    \hline
    PNAS (\cite{pnas}) & CIFAR10 & \textbf{3.2M} & 3.41 & 8 & 225\\
    \hline
    ENAS (\cite{enas}) & CIFAR-10 &  4.6M & 3.54 & 6 & \textbf{0.45}\\
    \hline
    DARTS (\cite{darts}) & CIFAR-10 &  4.6M & 2.76 & 7 & 4\\
    \hline
    NAO (\cite{nao}) & CIFAR-10 &  128M & \textbf{2.07} & 11 & 200\\
    \hline
    ProxylessNAS (\cite{plnas}) & CIFAR-10 & 5.7M & 2.08 & 6 & 8.3\\
    \hline
    SNAS (\cite{snas}) & CIFAR-10 & 2.85M & 2.8 & 7 & 1.5\\
    \hline
    Graph Hypernetworks (\cite{ghypernetworks}) & CIFAR-10 & 2.84M & 5.7 & 8 & 0.84\\
    \hline
    NASIB & CIFAR10 & 6.71M & 3.57 & \textbf{81} & 1.5\\
    \specialrule{.1em}{.05em}{.05em}
\end{tabular}
\label{tab:my_label}
\end{table}
\subsection{CIFAR-10}
For benchmarking CIFAR-10 experiments, we use a $12$ layer network. Between these $12$ layers, we insert $3$ factorized reduction blocks which perform downsampling of the image used and increase channel width. Each factorized reduction block is composed of one max-pooling filter and a $2\times2$ convolution filter with stride 2 which reduces both the spatial dimensions by a factor of $2$ and the output of both blocks is concatenated along the channel dimension, increasing the overall number of channels by a factor of $2$. We compare our result with other widely known NAS methods in Table~\ref{tab:my_label}.
We use the Resnet (\cite{resnet}) architecture as the backbone of our architecture space. Once the architecture search is completed, we obtain the final architecture as described in the section~\ref{method} which is retrained from scratch. We sample 5000 images randomly from the training data for the training of the architecture parameters. We use SGD with momentum of 0.9 for training of the architecture and use cosine update rule for learning rate update with initial learning set as 0.01. We also use some of the standard methods and modules used in CNN architectures like gradient clipping, cutout, L2-regularization, and batch normalization at every layer. Note that all these hyperparameters and additional modules remain unchanged for the training of the final compact architecture.
\section{Discussion}
\label{sec:discussion}
{\bf Filter Choices}: It is natural to question the validity of a big search space itself and focus on a curated search space e.g. in the case of CNN architecture search, this would mean only looking at the filters which have been used before in the well performing architectures. We were surprised by the discovery of a lot of unconventional filters like non-square and even sized ones. In order to confirm our hypothesis, we perform experiments by increasing the search space of well known architecture search method ENAS (\cite{enas}) and it also shows similar results as shown in Figure~\ref{enasvsbigenas}.
Furthermore, freeing up architecture search from bias would make it more applicable in domains beyond image recognition on CIFAR10. Using such a representative search space for image classification could prove beneficial, especially for datasets obtained from unconventional sources, for example, malware images \cite{malimg} - a dataset having only sequential correlations may discover filters of shape $n\times1$. Hence, it is better for the algorithm to rule out the unnecessary filters on its own instead of an expert designer. As previously discussed in section~\ref{cnnsearch}, in some cases it makes more sense to provide expert bias and hence our method also allows to have those biases in structural decisions in the form of priors over architecture parameters.

{\bf Efficiency Analysis}:
The main purpose of this work is not to introduce yet another new efficient algorithm. The design of the architecture search method and the size of the search space are the two factors which significantly contribute to the efficiency. Fig.~\ref{searchcomparison} sheds light on the relative efficiency of our method both on small as well as big search space. The search space induced by superkernels makes our method slightly inefficient compared to other methods. However, if the user has domain expertise, then they can assign higher probability to the desired candidate operations under our framework. This work adds another dimension to the efficiency of NAS methods which is efficient utilization of computational budget. NASIB optimizes computation budget utilization by starting with a huge search space induced by super kernels and adaptively trading off between exploration vs exploitation.

\iffalse
\textcolor{blue}{Whereas, the positive side of the proposed work is the introduction of another degree of freedom in terms of the computational budget for the neural architecture search.}

0. We present the problem of performing architecture search as a trade-off between architectures 
1. Why can't we decay lambda instead of having two policies? compare with SNAS also and mention that firstly after certain number of epochs it will simply become zero and only one element will be sampled every time for rest of the epochs
2. How is benchmarking on numbers is not fair? ex. proxyless nas uses pyramidnet as base network, mnasnet uses mobilenetv2.
4. How effective is path binarization (plnas, snas) and reduced sampling (plnas) for every iteration compared to all path open (darts), do some experiment for this one
5. Why to choose only one/two paths out of all (darts,plnas,snas), again some experiment is required.
6. Discovered architecture on colon cancer dataset and possible explanation behind the discovered kernels and their comparison with cifar10 discoverd kernels.
7. How our method will show similar performance to darts and etc. when provided with same environment and search space
8. Due to small lambda, training might be biased to initial results in a very low resource environment.
\fi
\section{Conclusion}
\label{sec:conclusion}
\iffalse
1. Contribution
2. Numbers and small comparison
3. What possible impact can it have
4. Future directions -
   - Improving the performance further
   - Applying this strategy on different datasets and novel tasks
\fi
In this paper, we present the problem of neural architecture search as a trade-off between the performance of the architecture search and computation resources available for performing the search. We propose a gradient based architecture search scheme which works well for varying computational budget scenarios and performs search efficiently over an augmented search space. Our experiments show the efficacy of the proposed method in adapting to the computation resources available and dynamically reducing the search space. We also present the intriguing results on the convolution filters discovered by bigger search spaces. We believe that our approach could spark off new research in this field (of NAS).
There are different ways in which this work can be extended. 
Combining the superkernel based search space with the multi-objective search could potentially discover interesting architectures.
This work can also be extended to search for the cell based architectures like RNNs and micro search space for CNNs. To foster reproducibility and usage of this method, we open source our implementation and all experiment configuration files at github.com/redacted\_for\_anonymity.

\bibliography{main}
\bibliographystyle{icml2019}

\end{document}